\documentclass{article}
\usepackage{amsmath,graphicx,mlspconf}

\toappear{2024 IEEE International Workshop on Machine Learning for Signal Processing, Sept.\ 22--25, 2024, London, UK}

\usepackage{cite}
\usepackage{amssymb,amsfonts}
\usepackage{algorithmic}
\usepackage{textcomp}
\usepackage{xcolor}
\usepackage{url}
\usepackage{booktabs}
\usepackage{multirow}
\usepackage{pifont}
\def\BibTeX{{\rm B\kern-.05em{\sc i\kern-.025em b}\kern-.08em
    T\kern-.1667em\lower.7ex\hbox{E}\kern-.125emX}}

\def\W{{\mathbf W}}
\def\x{{\mathbf x}}
\def\ii{{\hat{\imath}}}	
\def\ij{{\hat{\jmath}}} 
\def\ik{{\hat{\kappa}}}	

\def\bQ{{\mathbb Q}}

\title{Hierarchical Hypercomplex Network for\\ Multimodal Emotion Recognition}

\name{Eleonora Lopez, Aurelio Uncini and Danilo Comminiello\thanks{This work was supported by the Italian Ministry of University and Research (MUR) within the PRIN 2022 Program for the project ``EXEGETE: Explainable Generative Deep Learning Methods for Medical Signal and Image Processing", under grant number 2022ENK9LS, CUP B53D23013030006, and also by the European Union under the National Plan for Complementary Investments to the Italian National Recovery and Resilience Plan (NRRP) of NextGenerationEU,  Project PNC 0000001 D3 4 Health - SPOKE 1: Clinical use cases and new models of care supported by AI/E-Health based solutions, and Project “Future Artificial Intelligence Research” (PE0000013 - FAIR - Spoke 5: High Quality AI).}}
\address{Dept. Information Engineering, Electronics and Telecommunications (DIET), Sapienza University of Rome, Italy\\
Email: eleonora.lopez@uniroma1.it.}

\begin{document}

\maketitle

\begin{abstract}
Emotion recognition is relevant in various domains, ranging from healthcare to human-computer interaction. Physiological signals, being beyond voluntary control, offer reliable information for this purpose, unlike speech and facial expressions which can be controlled at will. They reflect genuine emotional responses, devoid of conscious manipulation, thereby enhancing the credibility of emotion recognition systems. Nonetheless, multimodal emotion recognition with deep learning models remains a relatively unexplored field. In this paper, we introduce a fully hypercomplex network with a hierarchical learning structure to fully capture correlations. Specifically, at the encoder level, the model learns intra-modal relations among the different channels of each input signal. Then, a hypercomplex fusion module learns inter-modal relations among the embeddings of the different modalities. The main novelty is in exploiting intra-modal relations by endowing the encoders with parameterized hypercomplex convolutions (PHCs) that thanks to hypercomplex algebra can capture inter-channel interactions within single modalities. Instead, the fusion module comprises parameterized hypercomplex multiplications (PHMs) that can model inter-modal correlations. The proposed architecture surpasses state-of-the-art models on the MAHNOB-HCI dataset for emotion recognition, specifically in classifying valence and arousal from electroencephalograms (EEGs) and peripheral physiological signals. The code of this study is available at \url{https://github.com/ispamm/MHyEEG}.
\end{abstract}


\section{Introduction}
\label{sec:intro}

Advancing the Brain-Comuter Interface (BCI) by understanding how the human brain encodes the world is at the core of neurocognitive research. A crucial part of this includes learning how emotions are related to brain signals such as electroencephalograms (EEGs). By deciphering the neural responses associated with different emotions, researchers can not only improve our understanding of human cognition but also pave the way for more sophisticated BCI systems capable of interpreting and responding to users' emotional states in real-time. However, emotions are intrinsically multi-modal, they are expressed through behavioral responses such as body language, facial expressions, and speech, as well as involuntary responses, which are reflected in physiological signals. Indeed, studies have shown that different emotions result in different responses of the brain which yield specific EEG signals \cite{li2021hierarchical}. Moreover, heart rhythm changes according to different emotions which can be detected through electrocardiograms (ECGs) \cite{hasnul2021electrocardiogram}. Galvanic skin response (GSR) provides a measure of the resistance of the skin, which decreases when one is experiencing emotions such as stress or surprise \cite{soleymani2011mahnob}. Finally, also the eyes can reveal insights into what emotions are being experienced, e.g. it has been shown that pupil diameter changes in different emotional states, increasing when feeling anger, fear, and anxiety or arousal and love \cite{lee2023emotion}. Given these relationships between physiological signals and emotional responses, researchers are starting to turn to the physiological approach for emotion recognition \cite{fu2022emotion}. Indeed, they are directly related to real emotions, unlike behavioral reactions, which can lead to systems that are prone to fake emotions and can be manipulated easily \cite{busso2004analysis}.

Nonetheless, existing works focus mostly on a single modality \cite{li2023brain, zhang2023attention} or rely on hand-crafted features extracted from the raw signals \cite{islam2021emotion}. Indeed, a multimodal approach yields a more powerful classifier since it takes into account the complementary information given by the different modalities, given that emotions are expressed in a multimodal way. Moreover, relying on handcrafted features requires extensive domain knowledge and represents a methodology rooted in the past. In contrast, contemporary deep learning encoders can autonomously acquire more discriminant features during the training. A recent work \cite{lopez2023hypercomplex}, addresses this problem by conducting a preliminary study and proposing a multimodal network equipped with a novel hypercomplex fusion module composed of parameterized hypercomplex multiplications (PHM). Parameterized hypercomplex neural networks (PHNNs) are models that operate in a hypercomplex domain which can be chosen directly by the user through a hyperparameter $n$. They represent a generalization of the more popular quaternion neural networks (QNNs) which are defined in the quaternion domain. Indeed, PHNNs can operate on any $n$-dimensional data, unlike quaternion models which work with quaternion data and are thus limited to 4-dimensional inputs. Moreover, they retain the advantages of QNNs, yielding lightweight models with the number of parameters reduced by $1/n$ and with the ability to model local relations among data dimensions \cite{grassucci2022phnns, zhang2021phm, lopez2024towards, lopez2024attention, lopez2022multi}. However, the model introduced in \cite{lopez2023hypercomplex}, HyperFuseNet, still presents some limitations. The utilization of hypercomplex algebra is confined solely to the fusion module and is not integrated into the encoder itself. Additionally, in general, these models have a tendency to overfit excessively, resulting in inadequate generalization performance. Therefore, in this paper, we build upon this preliminary study in order to address these problems. 

We propose a Hierarchical Hypercomplex (H2) model with a hierarchical structure wherein the encoders and the fusion module operate in the hypercomplex domain. The hierarchical structure allows to first learn embeddings of the single modalities exploiting relations among the channels that compose each modality through parameterized hypercomplex convolutions (PHCs). In contrast, the fusion module learns a fused representation from the embeddings of each modality, leveraging relations among the modalities themselves thanks to parameterized hypercomplex multiplications (PHMs). In detail, the encoders are tailored to each modality by setting the hyperparameter $n$ equal to the number of dimensions of each input signal. In this way, each encoder operates within its own distinct hypercomplex domain which is the natural domain of the specific signal. Thus, the encoders yield an enriched embedding thanks to convolutions and hypercomplex algebra properties. Indeed, convolutions are well-suited for processing multidimensional data, while hypercomplex algebra endows the encoders with the ability to model and exploit correlations among the various dimensions within each individual modality. Certainly, the modalities we employ, i.e., EEG, ECG, and eye data, are all multidimensional (except for GSR) and inherently exhibit correlations between their channels, an essential aspect of their information content. Thus, with this hierarchical design, we increase the performance of the state-of-the-art arousal and valence classification on the MAHNOB-HCI database \cite{soleymani2011mahnob}, achieving an F1-score of 0.557 and 0.685, respectively, which represents an improvement of 40.20\% and 57.11\%.

The rest of the paper is organized as follows. In section \ref{sec:related} we discuss the recent works on hypercomplex models and emotion recognition. In Section \ref{sec:method} we provide background on the theory behind hypercomplex networks and we describe the proposed architecture in detail. In Section \ref{sec:experiments} we discuss the experimental evaluation of our method. Finally, in Section \ref{sec:conclusion} we draw the conclusions.

\section{Related works}
\label{sec:related}

\begin{figure*}[t]
    \centering
    \includegraphics[width=0.7\textwidth]{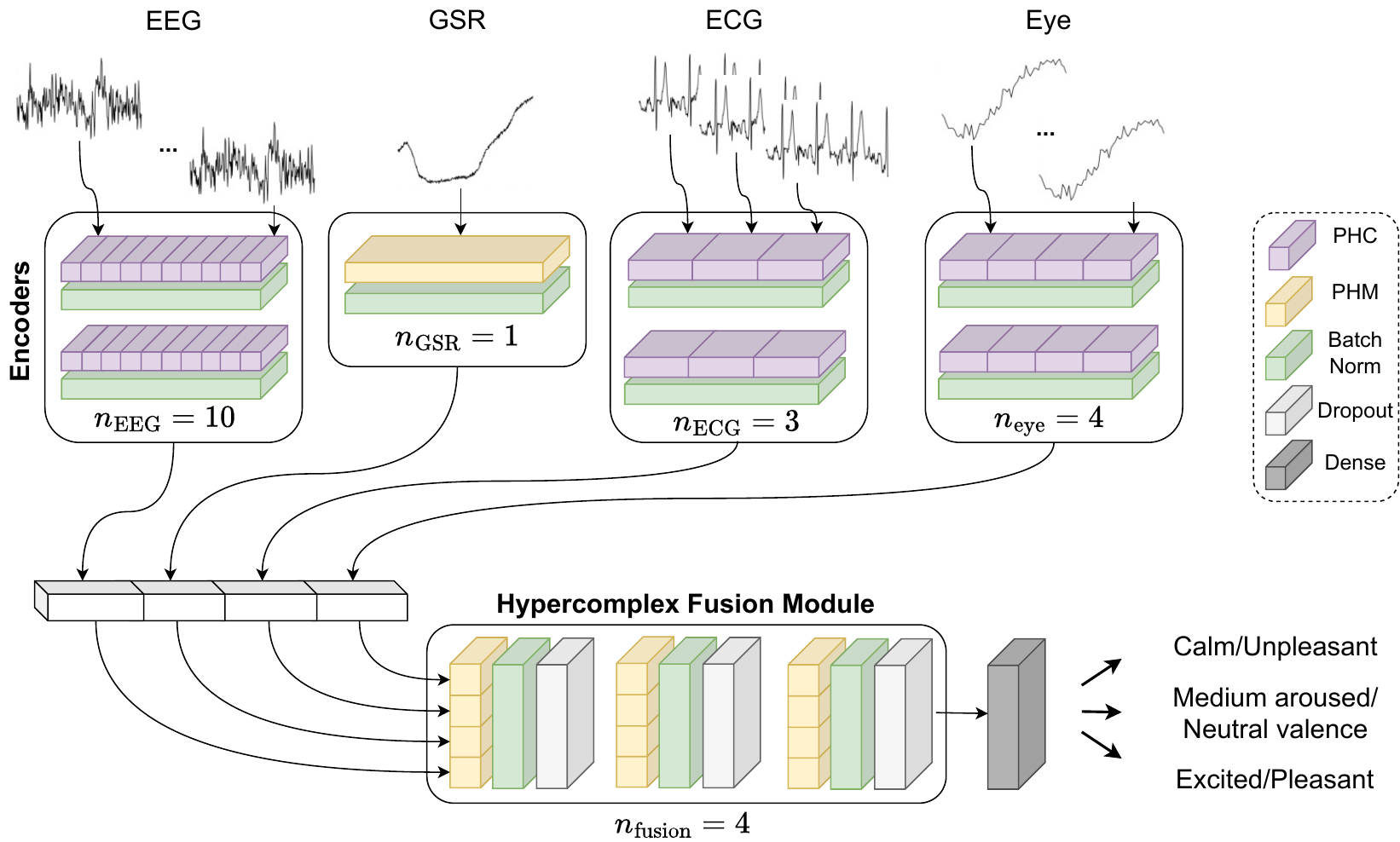}
    \caption{H2 model. The encoder learns enriched modality-specific embeddings by exploiting inter-channel relations within single modalities. The hypercomplex fusion module learns inter-modal correlations. Finally, a fully-connected layer produces the prediction for arousal/valence. In PHC and PHM layers, the smaller blocks indicate the number of submatrices of the weight matrix $\W$ following eq.\ref{eq:matirx}, which are shared among input dimensions and allow learning relations among them.}
    \label{fig:method}
\end{figure*}

Many studies have developed machine learning \cite{abdel2023efficient, liu2017realtime} and deep learning-based \cite{zhang2023attention, du2022efficient} systems for emotion recognition. Nevertheless, these methods require either extensive domain knowledge for extracting handcrafted features or employ popular extracted features such as differential entropy (DE) or power spectral density (PSD). These approaches do not allow the neural model to learn features on its own and require additional preprocessing steps \cite{islam2021emotion}. For these reasons, methods that learn directly from raw signals have started to emerge. A recent study has proposed a reinforcement learning approach inspired by brain emotion perception and shows the advantages of using raw signals \cite{li2023brain}. Furthermore, another approach to increase the performance of an emotion recognition system involves designing a multimodal framework \cite{islam2021emotion}. Indeed, recently proposed methods include a manifold learning-based technique for multimodal emotion recognition \cite{zhang2022multimodal}, a method to improve generalization across unseen target domains from EEG and eye movement signals \cite{gong2024ciabl}, and a hypercomplex-based architecture, HyperFuseNet, with a novel fusion module \cite{lopez2023hypercomplex}. In this paper, we tackle the limitations of handcrafted features and single-modality approaches by extending the method proposed in a preliminary study \cite{lopez2023hypercomplex}, HyperFuseNet. Although this model learns directly from raw multimodal signals, it suffers from poor generalization ability. To overcome this limitation, we introduce a hierarchical design, resulting in substantial improvements, as demonstrated by the experimental results in Section~\ref{sec:experiments}.

\section{Method}
\label{sec:method}

\subsection{Parameterized Hypercomplex Networks}

Parameterized hypercomplex neural networks (PHNNs) have been proposed in order to overcome the limitations of the popular quaternion models while keeping their advantages and useful properties \cite{zhang2021phm, grassucci2022phnns}. Indeed, quaternion models are limited to 4-dimensional inputs as they work with quaternions, i.e., $q = q_0 + q_1 \ii + q_2 \ij + q_3 \ik$ where $q_i \in \mathbb R$ and $\ii, \ij, \ik \in \bQ$. Instead, PHNNs are flexible to work with any $n$-dimensional data, where $n \in \mathbb{N}$ is a hyperparameter that users can adjust based on their specific requirements. This flexibility is achieved through a specific construction of the weight matrix within the layers comprising the PHNNs. Classic fully-connected and convolutional layers are equivalent to parameterized hypercomplex multiplication (PHM) layers and parameterized hypercomplex convolutional (PHC) layers, respectively. PHM and PHC layers are defined as standard linear and convolutional layers as:

\begin{equation}
\begin{split}
    \mathbf{y}_{\text{PHM}} &= \text{PHM}(\x) = \W \cdot \x + \mathbf{b}_{\text{PHM}}, \\
    \mathbf{y}_{\text{PHC}} &= \text{PHC}(\x) = \W*\x + \mathbf{b}_{\text{PHC}},
\end{split}
\label{eq:phm_phc}
\end{equation}

\noindent where $\x$ is the input and $\mathbf{b}_{\text{PHM}}$ and $\mathbf{b}_{\text{PHC}}$ are the bias terms. The core of these layers is how the weight matrix $\W$ is constructed through Kronecker products:

\begin{equation}
    \W = \sum_{i=1}^n \mathbf{A}_i \otimes \mathbf{F}_i.
\label{eq:matirx}
\end{equation}

\begin{table*}[t]
\centering
\caption{Results on MAHNOB-HCI of the proposed network compared against state-of-the-art models on arousal and valence classification. Results in bold and underlined are the best and second best, respectively.}
\label{tab:results}
\begin{tabular}{@{}|l|cc|lc|@{}}
\toprule
\multicolumn{1}{|c|}{\multirow{2}{*}{Model}} & \multicolumn{2}{c|}{Arousal}                              & \multicolumn{2}{c|}{Valence}                              \\\cmidrule(l){2-5} 
\multicolumn{1}{|c|}{}                       & \multicolumn{1}{c|}{F1-score}          & Accuracy         & \multicolumn{1}{l|}{F1-score}          & Accuracy         \\ \midrule
Dolmans                                      & \multicolumn{1}{c|}{0.389 $\pm$ 0.011} & 40.90 $\pm$ 0.62 & \multicolumn{1}{l|}{0.383 $\pm$ 0.012} & 40.24 $\pm$ 1.04 \\
HyperFuseNet                                 & \multicolumn{1}{c|}{\underline{0.397} $\pm$ 0.018} & \underline{41.56} $\pm$ 1.33 & \multicolumn{1}{l|}{\underline{0.436} $\pm$ 0.022} & \underline{44.30} $\pm$ 2.01 \\
H2 (ours)                                    & \multicolumn{1}{c|}{\textbf{0.557} $\pm$ 0.011} & \textbf{56.91} $\pm$ 0.99 & \multicolumn{1}{l|}{\textbf{0.685} $\pm$ 0.015} & \textbf{67.87} $\pm$ 1.48 \\ \bottomrule
\end{tabular}
\end{table*}

\noindent In this equation, matrices $\mathbf{A}_i$ learn the algebra rules directly from the data, while matrices $\mathbf{F}_i$ represent standard learnable weights of linear layers and filters of convolutional layers. Finally, $n \in \mathbb{N}$ defines the number domain in which the model operates, e.g. for $n=4$ it is equivalent to a QNN. Thus, they do not rely on algebras that exist only for domains in accordance with the Cayley-Dickson system, i.e., only for $n=2^m$ where $m \in \mathbb{N}$, such as quaternions ($n=4$), octonions ($n=8$), and so on. Instead, by modeling interactions between imaginary units through learnable matrices $\mathbf{A}_i$, they can be employed also for domains in which an algebra structure does not yet exist, e.g., for $n=3, 5, \cdots$. Moreover, this leads to a reduction of parameters by $1/n$, yielding very efficient models. Finally, they have the ability to learn both global relations, i.e., intra-channel interactions, and local relations, i.e., inter-channel interactions, which real-valued networks tend to ignore \cite{comminiello2024demystifying}.

\subsection{Hierarchical Hypercomplex Model}

In this section, we describe the proposed Hierarchical Hypercomplex (H2) model for multimodal emotion recognition, depicted in Fig.~\ref{fig:method}. The architecture has a hierarchical structure where encoders operating in different hypercomplex domains learn modality-specific embeddings, while the hypercomplex fusion module learns a fused embedding. Mainly, the hierarchical structure refers to the level of relations being considered, i.e., intra-modality and inter-modality. The encoders model relations between channels within a single modality, thus they exploit intra-modality relations. In contrast, the hypercomplex fusion module exploits relations among the different modalities themselves, i.e., inter-modal interactions. 
In detail, the encoders of EEG, ECG, and eye signals comprise two PHC layers with two batch normalization (BN) layers and ReLU activation functions. Instead, being GSR a 1-dimensional signal, its encoder is composed of a single PHM layer together with a BN layer and ReLU. Then, a global average pooling operation is applied to get the final latent representation. For PHC and PHM layers, we set $n_{\text{EEG}}=10, n_{\text{ECG}}=3, n_{\text{eye}}=4$, and $n_{\text{GSR}}=1$, in accordance with the number of channels of each different signal. In this way, the encoders are endowed with the ability to learn not only intra-channel relations, as any standard network, but also inter-channel relations. Thus, they exploit the inherent correlations among channels of the single modalities which enrich the latent representation of each signal. Moreover, by setting the $n$ parameter equal to the number of channels of each input signal, each encoder operates in a different hypercomplex domain which is also the natural domain of each input signal.

The hypercomplex fusion module employed in this architecture is a new version of the one introduced in a preliminary work \cite{lopez2023hypercomplex}. Specifically, it is modified by removing one PHM layer and incorporating more dropout layers in order to reduce overfitting, which the HyperFuseNet architecture struggled with.

To conclude, the H2 model can be seen as an extension of the method proposed in \cite{lopez2023hypercomplex}, namely HyperFuseNet. The main difference lies in the encoders, which before were composed of standard fully-connected layers in the real domain, instead of convolutional layers in the hypercomplex domain. Therefore, HyperFuseNet did not have the hierarchical structure that we introduce in this paper, which is also the main advantage of the proposed network. Moreover, we revisited the overall structure by reducing the number of layers by 1 in both the encoders and the hypercomplex fusion module in order to make them more proportionate to the quantity and nature of the input data. Also, for this reason, we add dropout layers with a dropout rate of $0.5$ between each PHM layer of the hypercomplex fusion module. 

\begin{figure}[t]
    \centering
    \includegraphics[width=0.9\linewidth]{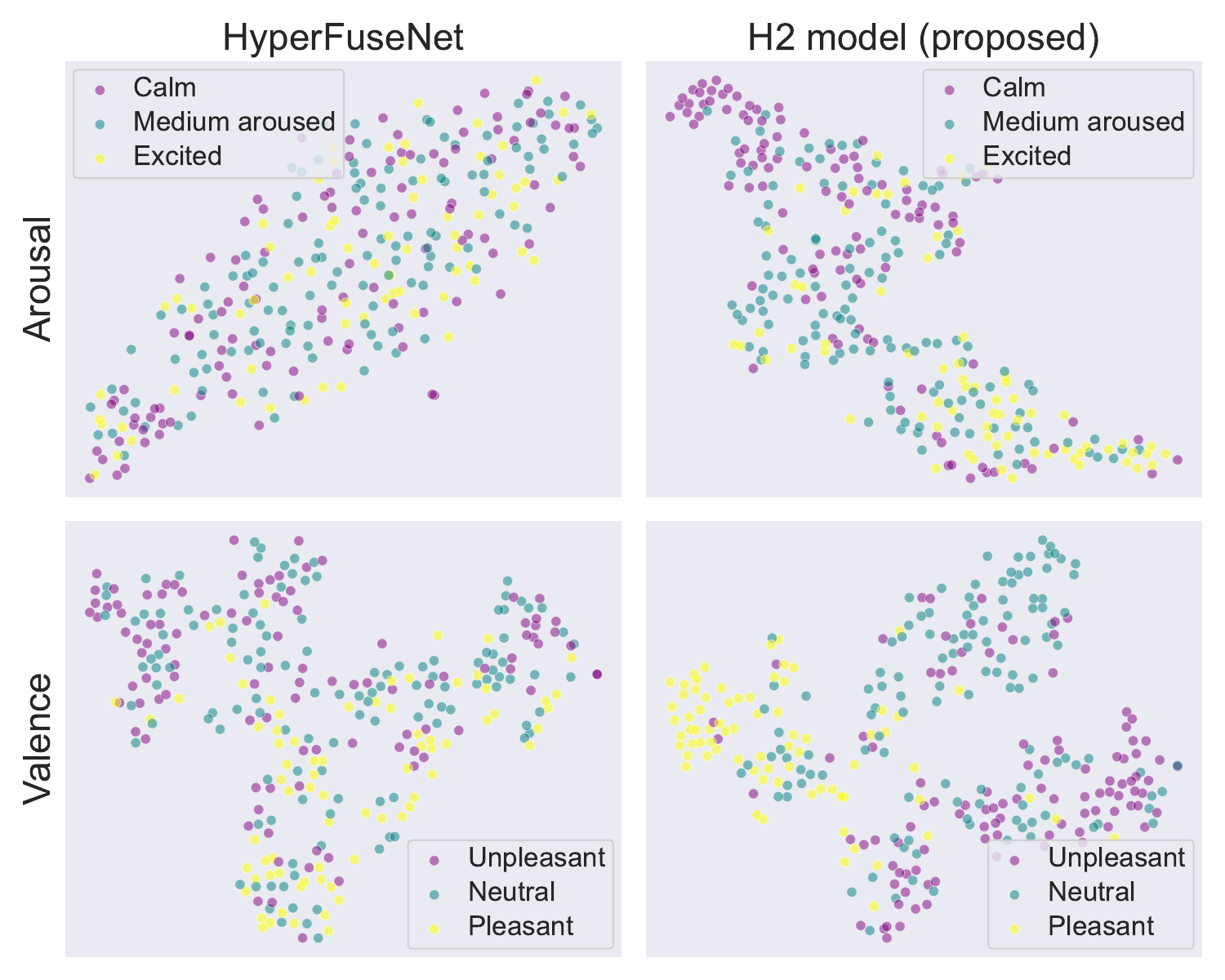}
    \caption{t-SNE feature visualization.}
    \label{fig:features}
\end{figure}

\section{Experimental results}
\label{sec:experiments}

\subsection{Dataset and preprocessing}

For training and evaluating our models, we utilize a publicly available database for affect recognition, MAHNOB-HCI \cite{soleymani2011mahnob}. It provides synchronized recordings of 27 participants during an experiment in which each subject was shown fragments of movies that induced different emotional responses. In detail, the subjects were monitored with video cameras, a head-worn microphone, an eye gaze tracker, as well as physiological sensors measuring EEG, ECG, respiration amplitude, and skin temperature. In this work, we employ as modalities the EEG, ECG, GSR, all recorded at 256Hz, and eye data, recorded at 60Hz. The latter comprises gaze coordinates, eye distances, and pupil dimensions. Each synchronized recording is labeled on a scale of arousal (calm, medium aroused, and excited) and valence (unpleasant, neutral, and pleasant).

We apply the same preprocessing as in \cite{lopez2023hypercomplex}. The dataset provides EEG with 32 electrodes, among which we select 10 most related to emotion, i.e., F3, F4, F5, F6, F7, F8, T7, T8, P7, and P8 \cite{topic2022reduced, msonda2021channel}. EEG, ECG, and GSR are first downsampled to 128Hz. Secondly, EEG signals are referenced to the average reference. Then, EEG and ECG are filtered with a band-pass filter of 1-45Hz and 0.5-45Hz, respectively, a low-pass filter for GSR at 60Hz, and a final notch filter at 50Hz for each of them. To account for the initial offset of GSR signals, baseline correction is performed by adjusting it relative to the mean value within the preceding 200ms of each trial. For the other signals, this correction is automatically achieved after the high-pass filters. Lastly, for eye data, we consider the average between measurements of the left and right eye, and we maintain -1 values as they indicate blinks or rapid movements which can be related to an emotional response. 

Samples are constructed by extracting segments of 10s from the original 30s of each trial. The dataset is split into training (80\%) and testing (20\%) in a stratified manner. The training data is augmented with scaling and Gaussian noise addition as in \cite{lopez2023hypercomplex}.

\subsection{Experimental setup}

For evaluating our models, we utilize as metrics the accuracy and the F1-score, i.e., the harmonic mean of the precision and recall, which accounts for imbalanced data.
The networks are trained with the Adam optimizer, cross-entropy loss, and the once-cycle policy with 0.425\% of increasing steps, linear annealing strategy, dividing factors of 10, maximum learning rate of $7.96 \times 10^{-6}$, and minimum and maximum momentum of 0.7403 and 0.8314, respectively. The number of epochs is set to 50, but we early stop the training when the F1-score does not improve for 10 epochs.


\subsection{Results and discussion}

The main results of our experiments are reported in Tab.~\ref{tab:results}. Our model is compared against two state-of-the-art architectures that operate with raw signals, i.e., HyperFuseNet \cite{lopez2023hypercomplex} and the multimodal model proposed by Dolmans et. al \cite{dolmans2021workload} which was originally designed for mental workload classification. The proposed hierarchical architecture significantly outperforms both HyperFuseNet and the other state-of-the-art model, as demonstrated by the results in Tab.~\ref{tab:results}. In both arousal and valence classification, our model brings a substantial improvement, i.e., of 40.20\% and 57.11\% for the F1-score, respectively. This great gap is due to several reasons. First, the other two models are quite prone to overfitting, thus when tested on unseen trials they are not able to generalize well. Instead, in our method we address these problems with additional dropout layers with a higher rate with respect to HyperFuseNet (which only had one dropout layer), reducing the overall number of layers by removing one in each encoder and fusion module, and, most importantly, integrating hypercomplex algebra also at encoder-level. Indeed, this last aspect allows to exploit both intra-modal and inter-modal relations in a hierarchical manner. This is the main advantage of the proposed model as we demonstrate in the ablation studies in the following section. In fact, it is not due to just a reduction of parameters given by hypercomplex algebra as this could result in underfitting, as we show in Section~\ref{sec:ablations}. Instead, thanks to its properties the encoders are able to leverage correlations among channels of single modalities, thus significantly outperforming HyperFuseNet. Moreover, Figure~\ref{fig:features} depicts the t-SNE visualization \cite{van2008tsne} of the features learned from H2 and HyperFuseNet. From the image, it is clear that HyperFuseNet struggles to learn discriminant representations. In contrast, in the features learned by the H2 model clusters start to become discernible, although there is room for further improvement.

\subsection{Ablation studies}
\label{sec:ablations}

\begin{table}[t]
\centering
\caption{Ablations on arousal. In parentheses we report the improvement percentage with respect to the best model so far according to the F1-score, starting from the top row.}
\resizebox{\linewidth}{!}{%
\begin{tabular}{@{}|l|c|c|c|@{}}
\toprule
Encoder                 & Params & F1-score       & Accuracy       \\ \midrule
HyperFuseNet            & 19.7M  & 0.396          & 41.03          \\ \midrule
Linear layers (less) & 16.2M  & 0.405 (2.0\%$\uparrow$)  & 42.11 (2.6\%$\uparrow$) \\
PHM layers              & 2.8M   & 0.393 (3.0\%$\downarrow$) & 44.41 (5.5\%$\uparrow$) \\
Conv. layers            & 9.2M   & \underline{0.425} (4.9\% $\uparrow$) & \underline{42.76} (1.5\%$\uparrow$) \\
PHC layers (H2)   & 2.4M   & \textbf{0.554} (30.4\% $\uparrow)$ & \textbf{56.91} (33.1\%$\uparrow$)\\ \bottomrule
\end{tabular}%
}
\label{tab:ablations_arousal}
\end{table}


\begin{table}[t]
\centering
\caption{Ablations on valence. In parentheses we report the improvement percentage with respect to the best model so far according to the F1-score, starting from the top row.}
\resizebox{\linewidth}{!}{%
\begin{tabular}{@{}|l|c|c|c|@{}}
\toprule
Encoder                 & Params & F1-score       & Accuracy       \\ \midrule
HyperFuseNet            & 19.7M  & 0.432          & 44.53          \\ \midrule
Linear layers (less) & 16.2M  & \underline{0.538} (24.5\%$\uparrow$) & \underline{53.95} (21.2\%$\uparrow$) \\
PHM layers              & 2.8M   & 0.511 (5.0\%$\downarrow$)   & 51.32 (4.9\%$\downarrow$) \\
Conv. layers            & 9.2M   & 0.521 (3.2\%$\downarrow$)   & 52.30 (3.1\%$\downarrow$)        \\
PHC layers (H2)         & 2.4M   & \textbf{0.684} (27.1\%$\uparrow$)& \textbf{67.76} (25.6\%$\uparrow$)\\ \bottomrule
\end{tabular}%
}
\label{tab:ablations_valence}
\end{table}

In this section, we study the impact of each added component of the proposed network H2 and we report the results in Tab.~\ref{tab:ablations_arousal} for arousal and in Tab.~\ref{tab:ablations_valence} for valence. In the first line of both tables, we show the results obtained with HyperFuseNet for reference. We investigate first the impact of reducing the number of layers of the encoders and fusion module, as described in Section~\ref{sec:method}. This simple tweak already improves performance resulting in slightly less overfitting. Then, we substitute fully connected layers with their hypercomplex counterpart, PHM layers with $n_{\text{m}}$, $\text{m} \in \{\text{EEG}, \text{GSR}, \text{ECG}, \text{eye}\}$ set as described in Section~\ref{sec:method}. In this scenario, the significant reduction in the number of parameters results in underperformance compared to using standard fully-connected layers. This also suggests that fully connected layers may not be optimal for this stage of the architecture in which the encoder should learn a modality-specific representation from multidimensional signals. Therefore, we substitute these with standard convolutional layers in the real domain. This allows the model to overfit less with respect to linear layers and improves the performance for arousal classification. For valence classification, the models are slightly less prone to overfit thus the performance in this scenario is comparable to using fully-connected layers. Finally, we introduce hypercomplex algebra, thus integrating convolutions in the hypercomplex domain, i.e., PHC layers, resulting in our final proposed H2 model with the hierarchical learning structure. From the results, it is clear that this is the crucial component that allows to achieve such a great improvement. Indeed, the parameters are reduced even more with respect to PHM layers in the encoder, however, the combination of convolution operations and hypercomplex algebra results to be optimal. In this scenario, the number of parameters, 2.4 million, is more appropriate for the task and data with respect to HyperFuseNet. However, the reduction of parameters is not the main advantage, as also the model version with PHM layers had the same size but resulted in being suboptimal. Instead, what allows the H2 network to generalize well is the ability of its encoders to learn embeddings enriched with relations among channels of the different signals thanks to hypercomplex algebra.


\section{Conclusion}
\label{sec:conclusion}

In this work, we proposed a multimodal hierarchical fully-hypercomplex model for emotion recognition from EEG and peripheral physiological signals. Specifically, the network is equipped with a hierarchical learning structure, where encoders learn intra-modal correlations, i.e., they exploit both intra-channel and inter-channel relations of each signal, and the hypercomplex fusion module learns inter-modal relations. Our main contribution is the introduction of hypercomplex algebra also at the encoder level with PHC operations, which results in a hierarchical structure. We found that this simple step allows to achieve much better results than previous state-of-the-art methods. Certainly, while the hypercomplex fusion module leverages inter-modal relations, the hypercomplex encoders exploit intra-signal correlations which results in enriched latent representations and a final improvement of 40.20\% and 57.11\% for the F1-score on arousal and valence classification.

\ninept
\bibliographystyle{IEEEtran}
\bibliography{biblio}

\end{document}